\documentclass[letterpaper, 10 pt, conference]{ieeeconf}  

\IEEEoverridecommandlockouts                              

\usepackage{amsmath} 
\usepackage{caption} 
\usepackage{placeins}

\usepackage{makecell}
\usepackage{algorithm, algorithmic}

\usepackage{booktabs}

\usepackage{graphicx}
\usepackage{subcaption}
\usepackage{bm}
\usepackage{multirow} 
\usepackage[table]{xcolor}
\usepackage{tabularx}
\usepackage{adjustbox}

\usepackage{siunitx}

\usepackage{bbm}
\usepackage{graphicx}
\usepackage{svg}
\usepackage{url}
\usepackage{hyperref}
\usepackage[T1]{fontenc}
\usepackage[flushleft]{threeparttable}
\usepackage[compact]{titlesec}

\titlespacing{\section}{0pt}{*1.2}{*1.0}

\usepackage{url}

\title{\LARGE \bf
Reinforcement Learning for Adaptive Planner Parameter Tuning: \qquad A Perspective on Hierarchical Architecture
}

\author{Wangtao Lu$^{1}$, Yufei Wei$^{1}$, Jiadong Xu$^{1}$, Wenhao Jia$^{2}$, Liang Li$^{1}$, Rong Xiong$^{1}$ and Yue Wang$^{1}$$^{\dagger}$
\thanks{This work was supported by the National Nature Science Foundation of China under Grant 62373322 and by Zhejiang Provincial Natural Science Foundation of China under Grant No. LD25F030001.}
\thanks{$^{1}$Wangtao Lu, Yufei Wei, Jiadong Xu, Liang Li, Rong Xiong, and Yue Wang are with the State Key Laboratory of Industrial Control Technology and Institute of Cyber-Systems and Control, Zhejiang University, Hangzhou, China.}
\thanks{$^{2}$Wenhao Jia are with the College of Information and Engineering, Zhejiang University of Technology, Hangzhou, 310023, China}
\thanks{$^\dagger$ Corresponding author, {\tt\small wangyue@iipc.zju.edu.cn}}
}

\begin{document}
\raggedbottom
\clearpage

\maketitle
\thispagestyle{empty}
\pagestyle{empty}

\begin{abstract}

 Automatic parameter tuning methods for planning algorithms, which integrate pipeline approaches with learning-based techniques, are regarded as promising due to their stability and capability to handle highly constrained environments. While existing parameter tuning methods have demonstrated considerable success, further performance improvements require a more structured approach. In this paper, we propose a hierarchical architecture for reinforcement learning-based parameter tuning. The architecture introduces a hierarchical structure with low-frequency parameter tuning, mid-frequency planning, and high-frequency control, enabling concurrent enhancement of both upper-layer parameter tuning and lower-layer control through iterative training. Experimental evaluations in both simulated and real-world environments show that our method surpasses existing parameter tuning approaches. Furthermore, our approach achieves first place in the Benchmark for Autonomous Robot Navigation (BARN) Challenge.
\end{abstract}

\section{INTRODUCTION}

Trajectory planning in highly constrained environments is a pivotal area of research within mobile robotics, with significant advancements made through both conventional methods and learning-based approaches \cite{quinlan1993elastic}, \cite{fox1997dynamic}, \cite{damanik2024lics}, \cite{wang2021agile}. In recent years, imitation learning (IL) has gained considerable momentum, particularly in the field of autonomous driving, where it is employed to generate trajectories or control commands based on demonstration data \cite{le2022survey}, \cite{chen2024end}. However, IL operates in an open-loop framework, where the actual execution of the robot is not factored into the learning process, making it particularly difficult to apply in highly constrained navigation tasks. On the other hand, reinforcement learning (RL) mitigates this issue by employing a closed-loop learning process that incorporates real-time feedback from the robot's interactions with the environment. Nevertheless, directly training velocity control policies via RL introduces additional difficulties, such as the need for extensive exploration and low sample efficiency.
 %

To mitigate these challenges, automatic parameter tuning systems for traditional planning algorithms have emerged as a promising solution \cite{xiao2022appl}. Planner parameters (e.g. maximum speed, inflation radius, etc.) can be interpreted as defining a parameterized space of feasible trajectories. Given the inherent guarantees offered by traditional planners, RL in the parameter space proves to be more efficient than learning directly in the trajectory or velocity control space, as it minimizes unnecessary exploration.
\begin{figure}[t]
  \centering
  \includegraphics[width=0.48\textwidth]{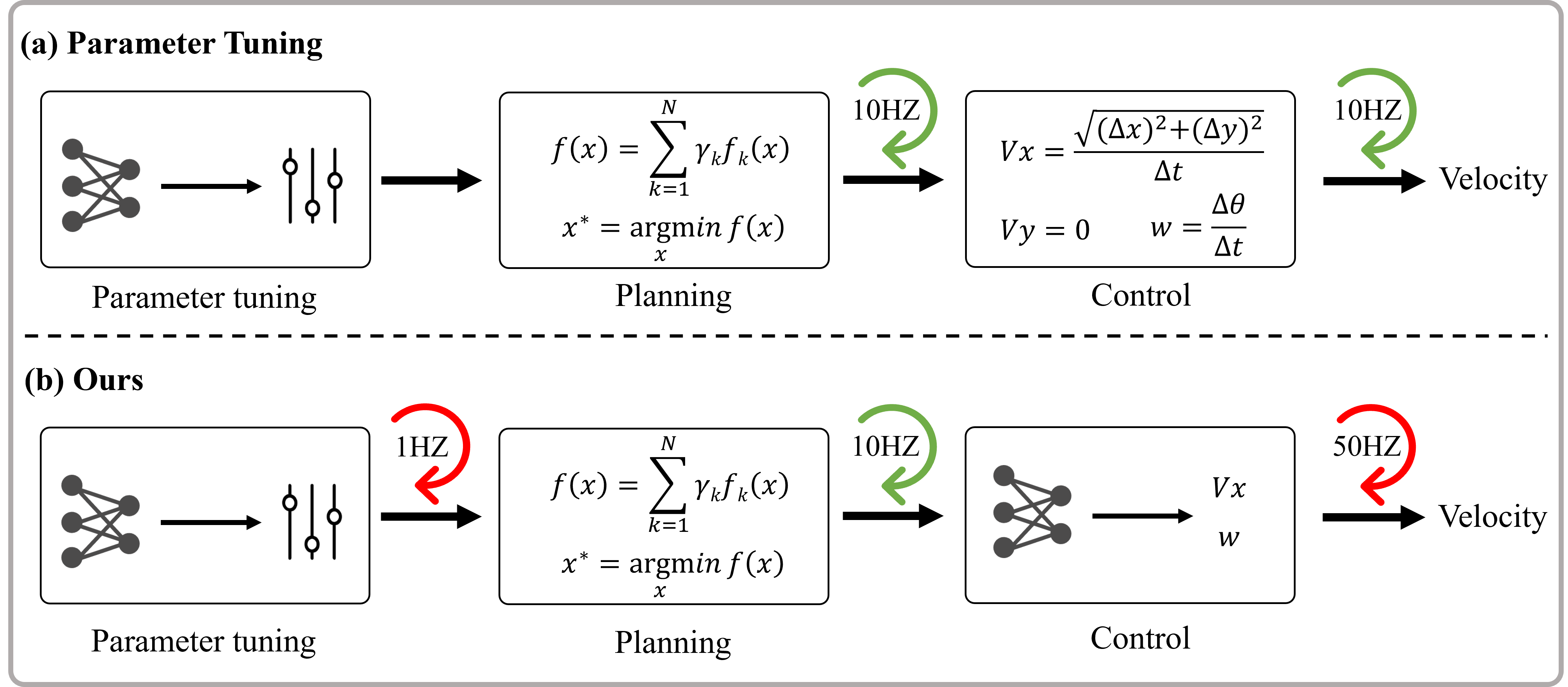} 
  \caption
  {\textcolor{black}Different frameworks for parameter tuning. (a): Existing framework. (b): Asynchronous hierarchical architecture.}
  \label{fig:introduce}
\end{figure} 
Current adaptive planner parameter tuning methods have demonstrated promising outcomes \cite{xu2021applr,wang2021apple,wang2021appli,xiao2020appld}. However, to further enhance performance, it is necessary to approach the issue from a broader perspective, considering the entire hierarchical framework of parameter tuning, planning, and control, rather than focusing solely on the parameter tuning layer.

In the closed-loop RL parameter tuning process, the high-level tuning network selects parameters based on the assumption that the controller will execute the planned trajectory as expected. However, as shown in Fig. \ref{fig:introduce} (a), due to the pure feedforward nature of the control system, there is often a discrepancy between the planned and executed trajectories, resulting in tracking errors. These tracking errors introduce noise into the RL closed-loop system, resulting in suboptimal parameter tuning decisions. They degrade the network's ability to learn and optimize the tuning policy effectively, as the errors cannot be corrected by the tuning network alone. Consequently, the performance of the tuning network is constrained by the limitations of the controller.

To overcome this challenge, we propose a hierarchical architecture consisting of low-frequency parameter tuning, mid-frequency planning, and high-frequency control, as shown in Fig. \ref{fig:introduce} (b). A policy operating at 1 Hz is trained to adaptively tune planner parameters, while the local planner, running at 10 Hz, generates trajectories and feedforward velocities for the low-level controller. To minimize tracking errors, we implement a feedback velocity controller operating at 50 Hz. Due to the challenges in accurately modeling the nonlinear disturbances in the system, we integrate an RL-based error compensator, which effectively mitigates these tracking errors. Additionally, LiDAR data is integrated into the controller’s input to enable adaptive obstacle avoidance. Finally, we propose an iterative training framework from the perspective of the hierarchical architecture, which alternates between optimizing the high-level parameter tuning and the low-level control, gradually improving the performance of both through repeated cycles.

Our contributions can be summarized as follows: 
\begin{itemize} 
\item We propose an alternating training method from the perspective of a hierarchical architecture, which integrates low-frequency parameter tuning, mid-frequency planning, and high-frequency control, enhancing both high-level parameter tuning and low-level control performance iteratively.
\item We propose an RL-based controller designed to reduce tracking errors while maintaining obstacle avoidance capabilities. 
\item We validate the effectiveness of the proposed method in simulation, achieving first place in the Benchmark for Autonomous Robot Navigation (BARN) challenge \cite{xu2023benchmarking}.
\item We conduct real-world experiments, demonstrating the method's sim-to-real transfer capability.
\end{itemize}

\section{RELATED WORKS}
\subsection{Parameter Tuning}
In traditional methods, a fixed set of parameters is often applied throughout the entire navigation task, resulting in suboptimal performance when environmental conditions change. Furthermore, parameter tuning typically requires expert knowledge and extensive trial-and-error experimentation \cite{zheng2021ros}. To mitigate this dependence on experts, techniques such as fuzzy logic \cite{teso2019predictive} and gradient descent \cite{bhardwaj2020differentiable} have been introduced to automatically identify suitable parameters for specific navigation scenarios.

The introduction of Adaptive Planner Parameter Learning (APPL) \cite{xiao2022appl} enables real-time parameter tuning, allowing robots to adaptively select optimal parameters in response to environmental changes during navigation. APPL methods can be broadly classified into two categories: IL-based methods and RL-based methods. Xiao $et \ al$. \cite{xiao2020appld} utilize human demonstrations to adjust parameters according to different navigation contexts. However, this approach is heavily dependent on the quality of the demonstrations and is limited to deployment within the training environment. Wang $et \ al$. \cite{wang2021appli} focus on learning through human interaction, intervening only when parameters are unsuitable, thereby helping the planner avoid local minima traps. Nonetheless, these IL-based methods require extensive human intervention data and are confined to discrete parameter adjustments.

Xu $et \ al$. \cite{xu2021applr} propose an RL-based approach to develop a continuous parameter selection policy, eliminating the need for demonstration data or human intervention, while offering a higher performance ceiling. However, designing an effective reward function to evaluate the quality of the current parameters remains a significant challenge. To address this, Wang, $et \ al$. \cite{wang2021apple} introduces human evaluations as dense rewards in RL, scoring planning results to accelerate training.

Current parameter tuning efforts predominantly focus on optimizing the tuning layer, often overlooking the limitations of the control layer. In contrast, we address the problem from a holistic perspective of the entire hierarchical architecture. By adopting different frequencies for each layer—low-frequency for parameter tuning, mid-frequency for planning, and high-frequency for control—we aim to iteratively improve overall system performance and reduce tracking errors that cannot be resolved through parameter tuning alone.

\subsection{End-to-end Navigation}
End-to-end navigation methods, which directly map raw sensor data to path points or motion commands, hold significant potential for achieving expert-level performance in robotic navigation \cite{muhammad2020deep} \cite{luo2024delving} \cite{chib2023recent}.


RL is a typical closed-loop learning approach, which fully incorporates the robot's execution feedback.
Recently, numerous studies have applied RL to address navigation challenges \cite{dong2023review} \cite{chemin2024study}. While many of these works focus on learning the complete navigation stack, research conducted by \cite{xiao2022motion} suggests that targeting navigation subsystems for learning can lead to enhanced outcomes. In \cite{xu2023benchmarking}, various RL-based navigation planning algorithms are evaluated using the BARN environment. Their benchmark reveals that traditional methods achieve the highest success rates in this context, while RL approaches show potential for the fastest task completion.

\begin{figure}[t]
\centering
\includegraphics[width=0.48\textwidth]{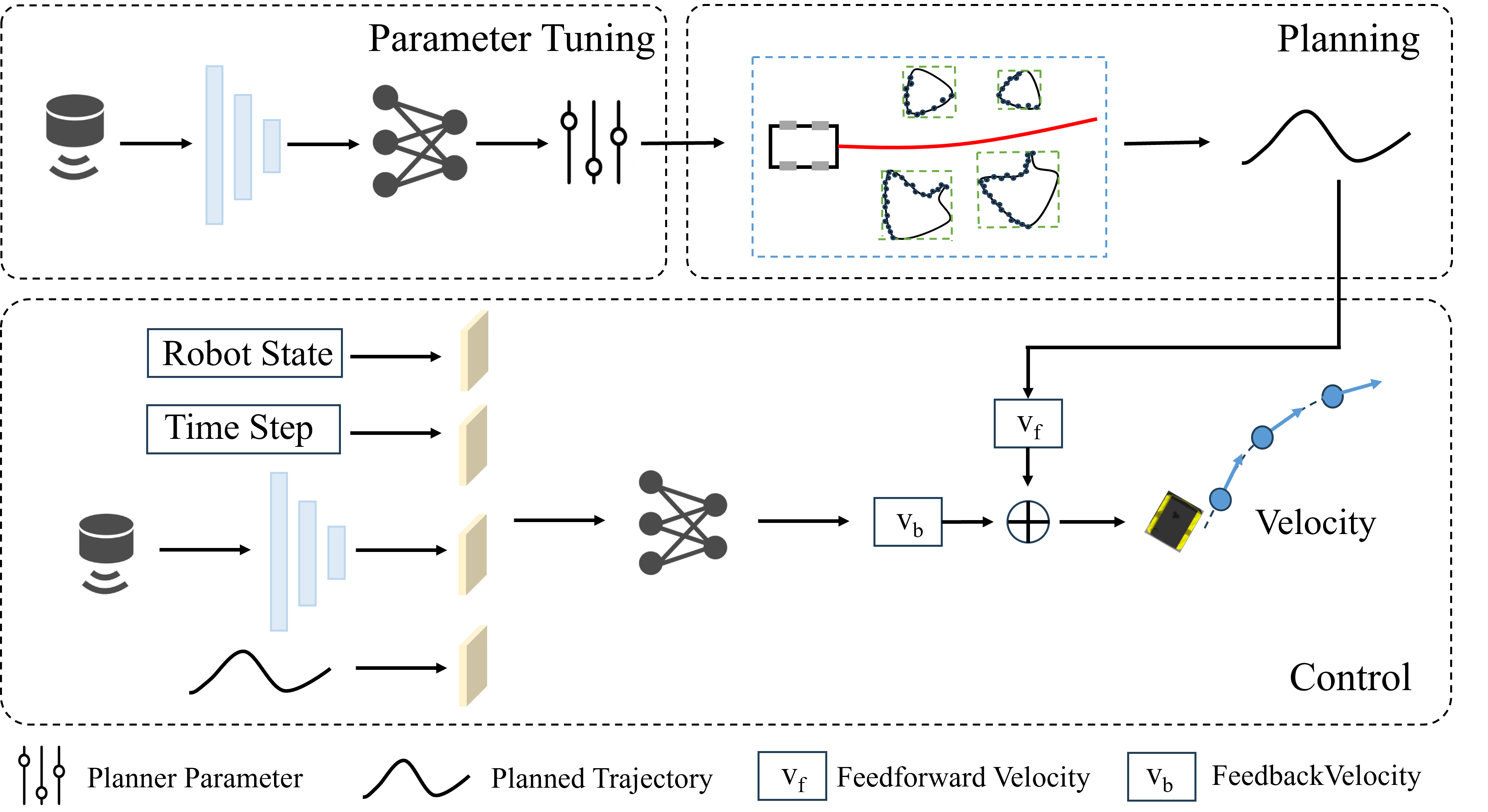}
\caption{Illustration of the hierarchical architecture. The parameter tuning network selects appropriate parameters for the local planner, which adjusts these parameters to generate both the trajectory and feedforward velocity. The controller then calculates the feedback velocity using LiDAR data, robot state, the planned trajectory and time step. The final speed control command is produced by combining the feedback and feedforward velocities.}
\label{fig:approach}
\end{figure}

Our approach integrates traditional methods with learning-based techniques, performing learning in both the parameter space and the velocity feedback space. By leveraging the robustness of traditional planning algorithms, our method enhances sample efficiency and ensures greater stability during the training process.

\section{APPROACH}

In this section, we present the proposed hierarchical architecture, as illustrated in Fig. \ref{fig:approach}. Both the parameter tuning and control strategies are trained iteratively, resulting in enhanced overall system performance.
\subsection{Problem Setup} 

\begin{figure}[t]
\centering
\includegraphics[width=0.46\textwidth]{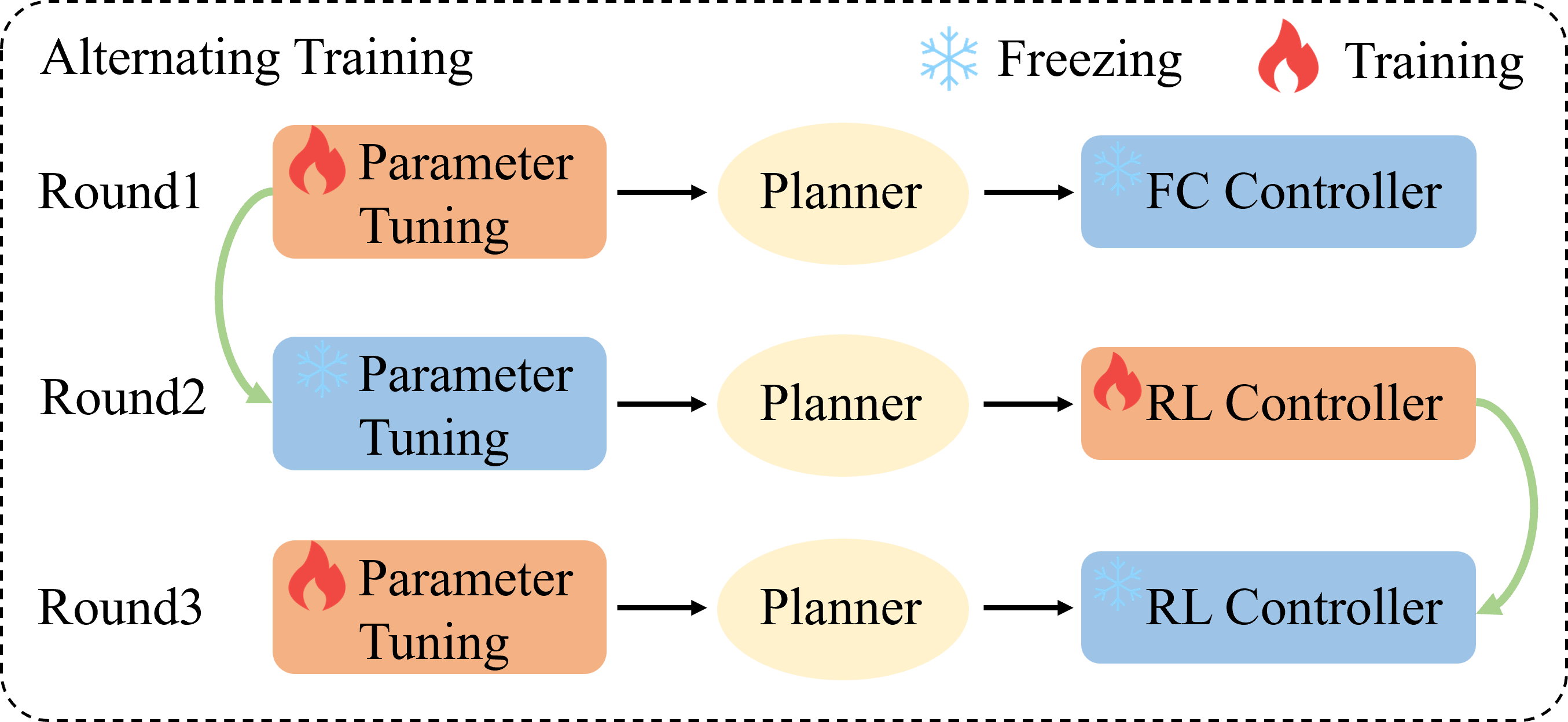}
\caption{Illustration of alternating training of the hierarchical architecture. FC Controller represents feedforward controller. RL Controller represents the proposed controller.}
\label{fig:al_training}
\end{figure}

We formulate the parameter tuning and control compensation problems as a Markov Decision Process (MDP), described by the tuple $(\mathcal{S}, \mathcal{A}, \mathcal{P}, \mathcal{R}, \gamma)$, where $\mathcal{S}$ is the state space, $\mathcal{A}$ is the action space, $\mathcal{P}$ is the state transition model, $\mathcal{R}$ is the reward function, and $\gamma$ is the discount factor. During each training phase, we fix one component while training the other. Specifically, when training the parameter tuning network, the controller remains fixed, and conversely, the parameter tuning network is fixed when training the controller, as shown in Fig. \ref{fig:al_training}. The following provides a detailed explanation of this training approach.

\subsection{Parameter Tuning}
\textbf{State Space:}
We define the state space exclusively by laser readings and use a variational auto-encoder (VAE) network \cite{kingma2013auto} to embed them as a local scene vector. 

\textbf{Action Space:}
The action is a set of hyperparameters of the local planner, such as $inflation  
 \ dist$, $maximum \ speed$, $weight \ obstacle$, etc. After tuning these parameters, the local planner generates a trajectory and feedforward velocity.

\textbf{Reward Function:}
To guide the policy optimization, we design a reward function that considers target arrival and collision avoidance:
\begin{equation}
    R = R_g + R_c + R_f
\end{equation}
Specifically, $R_f = -1$ is used as a penalty for each step before reaching the goal, $R_g$ rewards progress towards goal:
\begin{equation}
    R_g = 
    \begin{cases} 
    r_{\text{arrival}} & \text{if goal reached} \\
    w_1(d_{t-1} - d_t) & \text{otherwise}
    \end{cases}
\end{equation}
in which $w_1$ represents the weight, $d_t$ denotes the distance from the center of mass of the robot to the goal position.
$R_c$ is a binary penalty that is only active when collisions occur:
\begin{equation}
    R_c = 
    \begin{cases} 
    r_{\text{collision}} & \text{if collision} \\
    0 & \text{otherwise}
    \end{cases}
\end{equation}
\begin{figure}[t]
\centering
\includegraphics[width=0.46\textwidth]{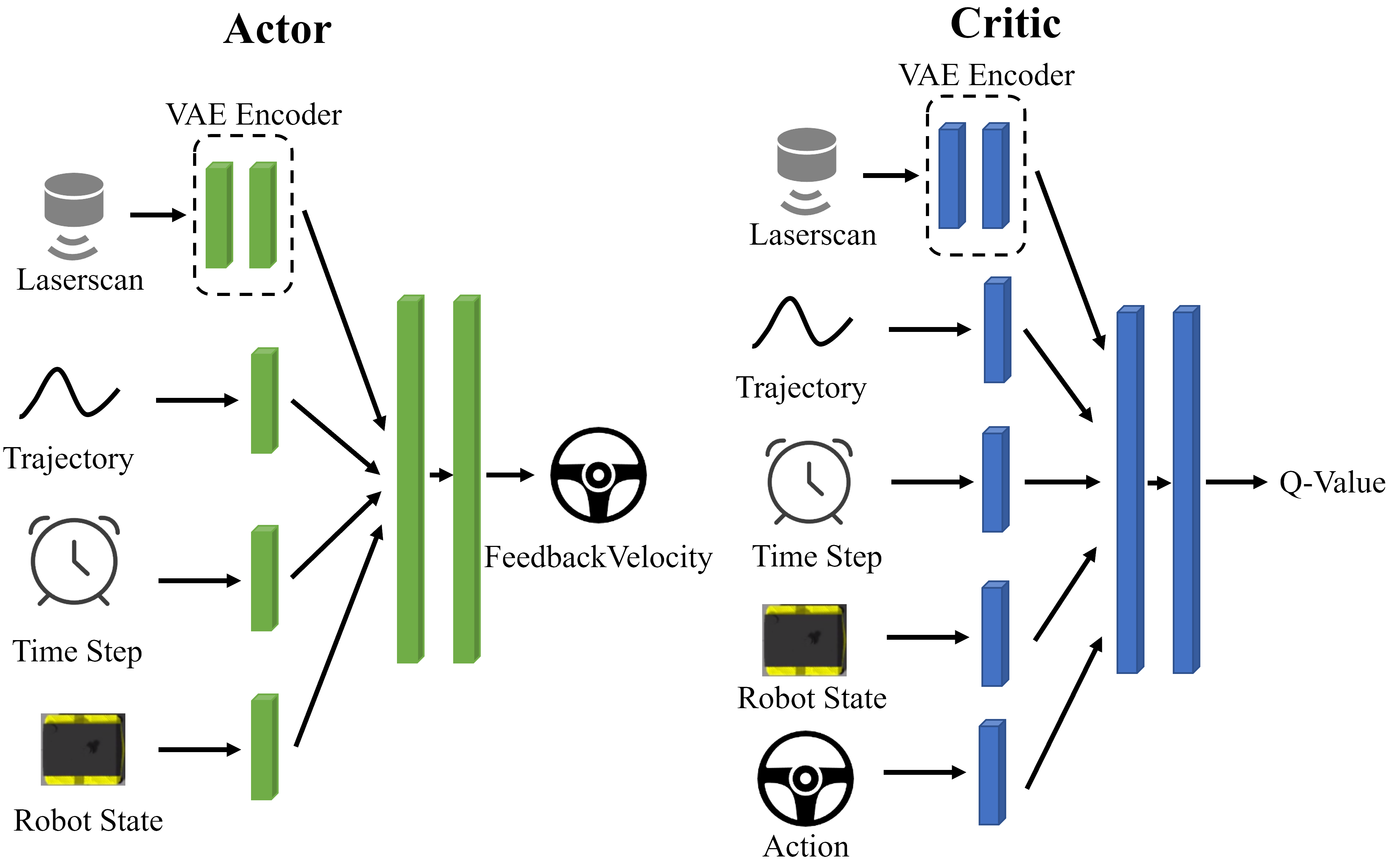}
\caption{The detailed architecture of the
 Actor-Critic network. The input laser scan data is encoded into a latent space using a VAE. The rest of the network consists of Multi-Layer Perceptron (MLP).}
\label{fig:actor-critic}
\vspace{-0.5cm}
\end{figure}

\subsection{Controller Design}
It is important to note that the planning frequency is set to 10 Hz, while the controller operates at 50 Hz. Consequently, the controller is formulated as a fixed-length MDP with an episode length of 5 steps, where each episode tracks the trajectory over a time horizon of 0.1 seconds (with each step covering 0.02 seconds). The planned trajectory is processed using linear interpolation to generate a series of equidistant poses. Incorporating both the time step and the processed trajectory is crucial to supporting the asynchronous nature of the design, as the controller must continuously adjust its actions based on real-time feedback. The controller’s primary objective is to minimize the tracking error between the predicted and actual poses during execution.

\textbf{State Space:}
The state space consists of laser readings, relative trajectory way points, time step, current relative robot pose (relative to the starting point of each episode) and robot's velocity. 

\textbf{Action Space:}
The action is a predicted value of the feedback velocity, which compensates for the tracking error during the process, and together with the feedforward velocity, forms the final velocity command.

\textbf{Reward Function:}
Our objective is to minimize tracking error and ensure collision avoidance, for which we design a reward function:
\begin{equation}
    R^{\prime} = R_t + R_{c^{\prime}}
\end{equation}
\begin{equation}
    R_t = w_2||p_t - p_*|| 
\end{equation}
in which $p_t$ denotes the current relative robot pose, $p_*$ represents the expected robot pose at specific waypoints along the relative trajectory, which are determined by querying the trajectory at the corresponding time steps.
$R_{c^{\prime}}$ is a binary penalty that is only active when collisions occur:
\begin{equation}
    R_{c^{\prime}} = 
    \begin{cases} 
    r_{\text{collision}}^{\prime} & \text{if collision} \\
    0 & \text{otherwise}
    \end{cases}
\end{equation}

\subsection{Alternating Training}
In this work, we employ alternating training to improve both high-level parameter tuning and low-level control performance. During alternating training, certain components of the system remain static (represented by parameter freezing), while others undergo active optimization, as shown in Fig. \ref{fig:al_training}. This approach promotes a balance between stability and adaptability, allowing different parts of the architecture to refine their parameters without destabilizing the entire system.

For the RL tasks, we utilize the Twin Delayed Deep Deterministic Policy Gradient (TD3) algorithm \cite{fujimoto2018addressing}, which is well-suited for robotic control in continuous action spaces. TD3 provides several advantages, including reduced bias, improved stability, and enhanced robustness, making it ideal for the hierarchical framework we propose. The controller's network architecture is shown in Fig. \ref{fig:actor-critic}, and the parameter tuning network follows a similar design.


\begin{figure}[t]
  \centering
  \includegraphics[width=0.46\textwidth]{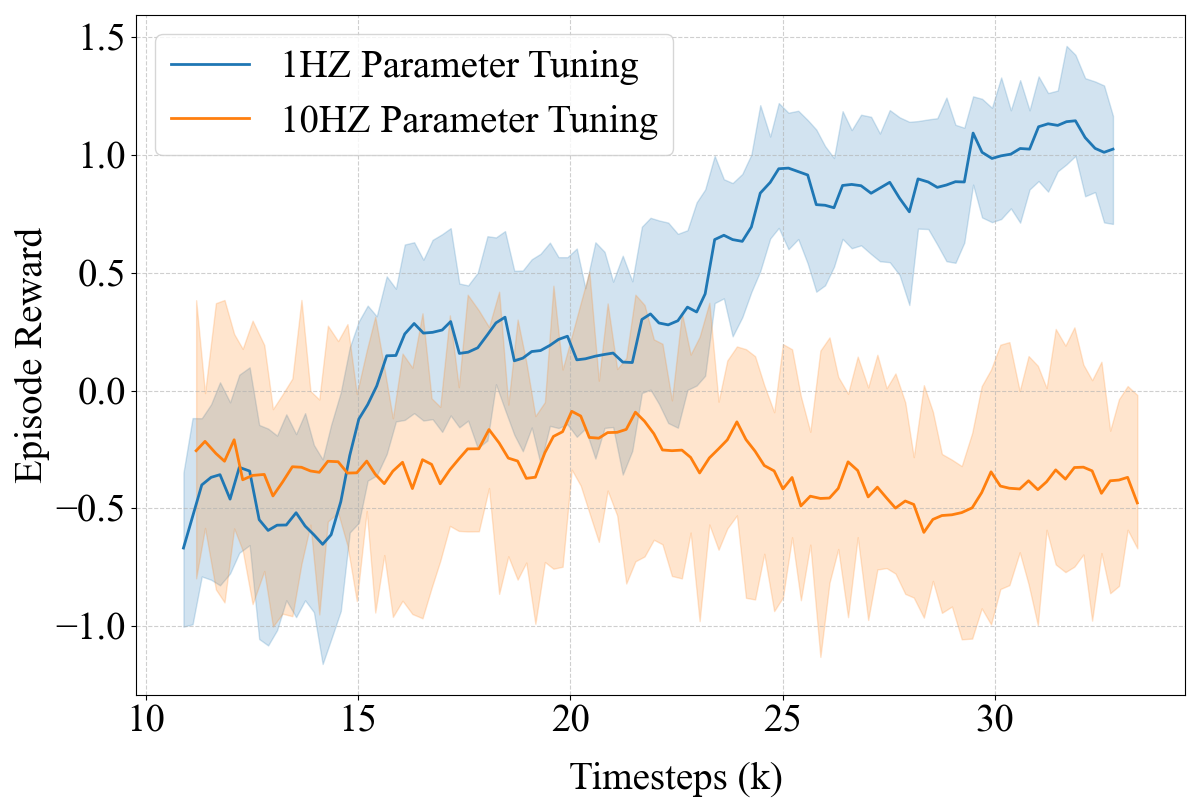} 
  \caption
  {Comparison of training loss curves. The local planner operates at 10 Hz, and we compare the training loss between 1 Hz and 10 Hz parameter tuning.}
  \label{fig:training efficiency}
  \vspace{-0.3em}
\end{figure} 
\begin{table}[tbp]
    \centering
    \caption{Performance analysis of the alternating training. PT represents parameter tuning, FC represents feedforward controller, RC represents RL-based controller.}
    \label{tab:comparison}
    \small 
    \begin{tabularx}{\linewidth}{@{}l*{4}{>{\centering\arraybackslash}X}@{}}
        \toprule
        Method       & PT \qquad Freq. (Hz)& Controller Freq. (Hz) & Success Rate (\%)↑ & Completion Time (s)↓ \\
        \midrule
        TEB+FC          & -        &  10Hz      &       70   &    26.7               \\
        PT+FC           &  1Hz          &    10Hz       & 84                  &     13.2                \\
        PT+RC   &      1Hz            &  50Hz         &   90                &            12.9         \\
        2PT+RC   &      1Hz            &   50Hz         &     \textbf{98}             &             \textbf{10.2}       \\
        \bottomrule
    \end{tabularx}
\end{table}

\section{EXPERIMENTS}
In this section, we analyze the impact of different frequencies within the three-layer hierarchical architecture, focusing on how the low-frequency parameter tuning, mid-frequency planning, and high-frequency control affect overall system performance. We also conduct ablation studies on the controller to highlight its effectiveness. Additionally, we perform comparative experiments in both simulation and real-world environments to demonstrate the robustness and efficiency of the proposed method.
\subsection{Implementation}
In this work, we utilize the BARN Challenge environment as the primary testbed for both training and evaluating the proposed algorithm. From the BARN dataset's 300 scenarios, arranged from simplest to most complex, we randomly select 50 environments as the test set, with the remaining serving as the training set. The experiments are conducted using a Jackal robot, equipped with a 270° field-of-view, 720-point 2D Velodyne LiDAR for precise environmental sensing, and capable of a maximum velocity of 2 m/s. We leverage the Robot Operating System (ROS) for system integration and communication, using the $move\_base$ package as the navigation stack. The global path is planned using the Dijkstra \cite{dijkstra2022note} algorithm, while local trajectory optimization is handled by the Timed Elastic Band (TEB) \cite{rosmann2012trajectory} planner. The parameter tuning module learns a policy to dynamically adjust TEB parameters, including $max\_vel\_x$, $max\_vel\_theta$, $weight\_obstacle$, and $inflation\_radius$. Dynamic tuning of these parameters is achieved through the ROS dynamic reconfigure client, enabling real-time adaptability.


\subsection{Impact of Frequency on Hierarchical Architecture}
In this experiment, we investigate the impact of different frequencies on the hierarchical architecture. Since the planner operates at a frequency of 10 Hz, we compare the performance of the tuning network running at 1 Hz and 10 Hz. Fig. \ref{fig:training efficiency} shows that operating the network at 1 Hz is more beneficial for policy learning. This is because the quality of the parameters can only be accurately assessed after executing the trajectory. If tuning and planning occur at the same frequency, only short-term trajectories are executed, which fails to adequately capture the impact of the parameters.

\begin{figure}[t]
  \centering
  \includegraphics[width=0.46\textwidth]{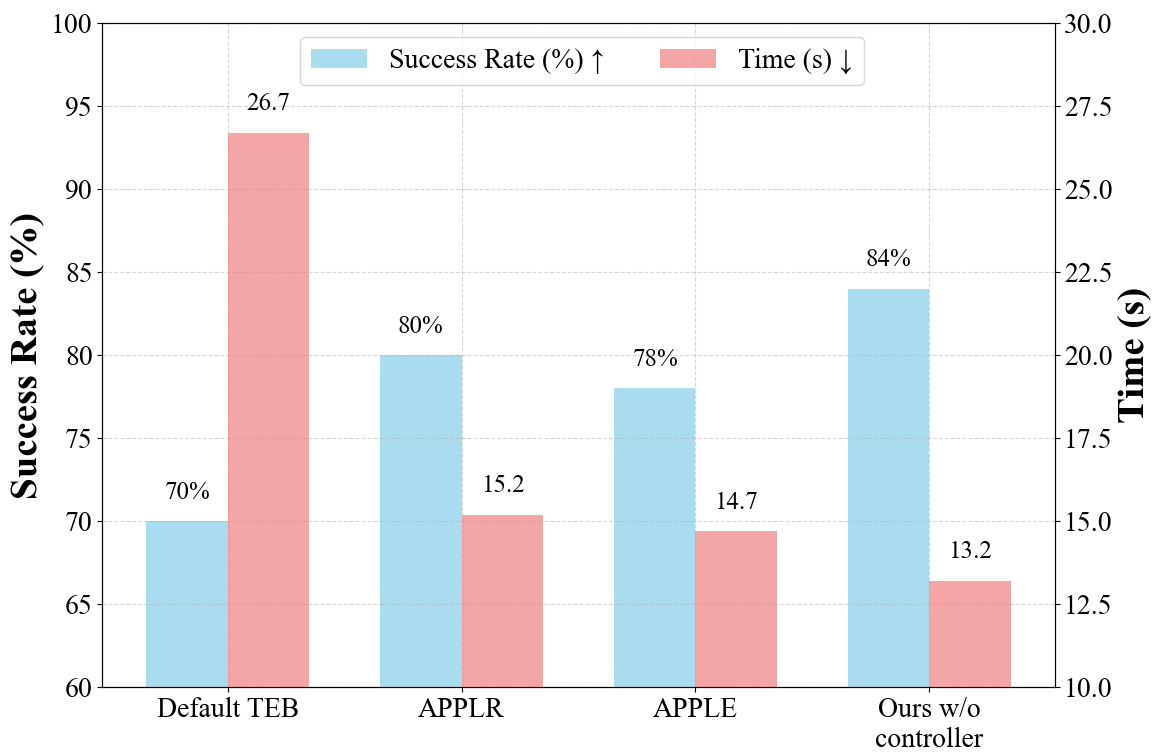} 
  \caption
  {Comparison of the parameter tuning method. Default TEB means running TEB without parameter tuning.}
  \label{fig:tuning_compare}
  \vspace{-0.3cm}
\end{figure} 
As shown in Table \ref{tab:comparison}, the comparison highlights the impact of varying frequencies in both the parameter tuning network and the controller, as well as the benefits of iterative training. In the baseline method (TEB), with a 10 Hz feedforward controller (FC) and no parameter tuning, the success rate is 70\%, and the task completion time is 26.7 seconds. When a parameter tuning network (PT) is introduced at 1 Hz with the controller still operating at 10 Hz, the success rate increases to 84\%, and the completion time significantly drops to 13.2 seconds. This improvement illustrates the advantage of automatic parameter tuning, which optimizes the planning process by adjusting the planner parameters.

Further performance gains are observed with the introduction of an RL-based controller running at 50 Hz (PT+RC). The success rate rises to 90\%, and the completion time improves to 12.9 seconds. The RL-based controller’s higher frequency enables more precise trajectory following, which helps reduce tracking errors and increase the success rate.

Finally, when an additional parameter tuning iteration (2PT+RC) is included alongside the RL-based controller, the success rate reaches 98\%, with a further reduction in completion time to 10.2 seconds. This demonstrates the value of iterative training, leading to more efficient task completion and higher overall success rates.



\subsection{Comparative Study of Parameter Tuning}
Our parameter tuning method differs from existing approaches in two key aspects: state input and reward function. While existing methods use raw laser scans as the state input \cite{xu2021applr}, we employ a pre-trained VAE as a state extraction network, using the extracted laser features as the state representation. This effectively serves as a dimensionality reduction technique. In terms of reward function design, we replace local progress with global progress, enabling the policy to focus more on overall task completion. Additionally, we remove the obstacle avoidance term from the reward function, delegating this task to the RL-based controller. The comparison results are shown in the Fig. \ref{fig:tuning_compare}.


\subsection{Ablation Study of Controller}
To validate the effectiveness of our proposed controller, we conduct a series of ablation studies. In these experiments, all parameters remain fixed after a single round of network tuning.

We systematically evaluate the impact of different controller output strategies. Specifically, we compare a full velocity output approach, where the controller directly determines the final execution velocity, with a feedback-only velocity output, where the controller provides feedback velocity that is combined with feedforward velocity to compute the final execution velocity.
\begin{figure}[t]
  \centering
  \includegraphics[width=0.46\textwidth]{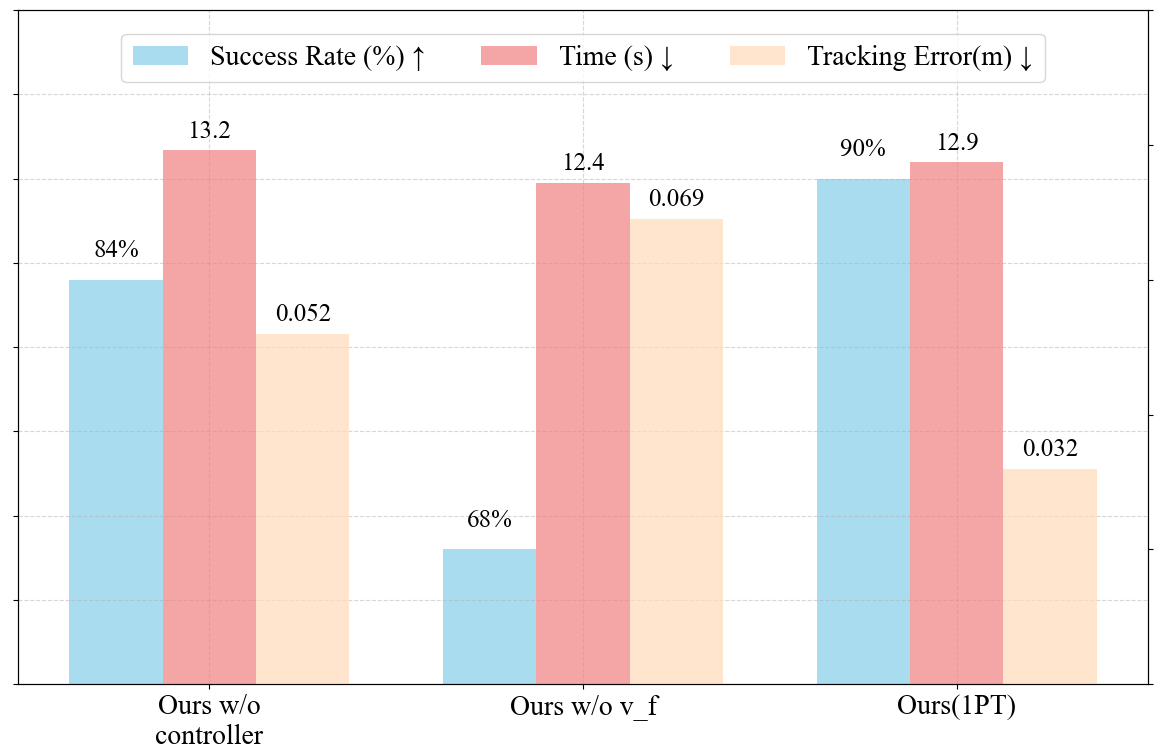} 
  \caption
  {Ablation study of controller. Ours w/o controller represents the feedforward controller, Ours w/o v\_f represents the controller output the full velocity, Ours(1PT) represents the proposed controller. All experiments are conducted after the parameter tuning network undergoes a round of training.}
  \label{fig:controller_ablation}
\end{figure} 

As shown in Fig. \ref{fig:controller_ablation}, the full velocity output approach proves challenging for the policy, as directly learning a control strategy requires managing both feedforward and feedback components simultaneously. This added complexity makes it more difficult for the policy to learn stable and consistent behavior. In contrast, when the controller outputs only feedback velocity, combining it with feedforward velocity, we observe significant improvements in performance. This is because the reduced complexity of the feedback-only strategy leads to more stable training, fewer unnecessary exploration steps, and improved robustness in task execution. 

Since TEB operates at 10 Hz, the tracking error is defined as the distance between the robot’s actual pose and the expected pose over a 0.1-second interval. In comparison, our proposed method effectively reduces tracking error while improving both the success rate and task completion speed.



\begin{figure}[t]
  \centering
  \includegraphics[width=0.46\textwidth]{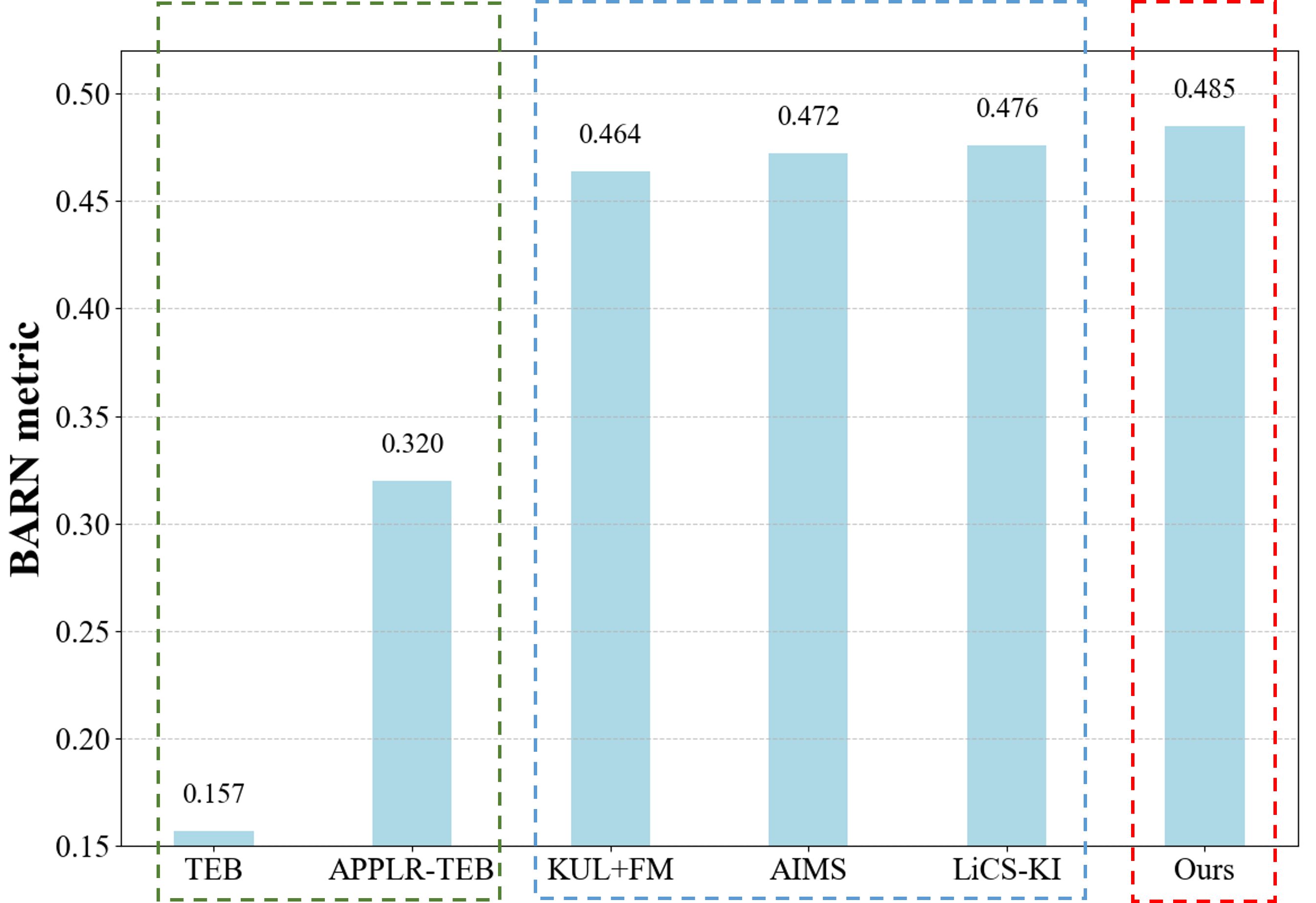} 
  \caption
  {The performance of autonomous navigation systems under different methods on the BARN dataset and 0.5 is the maximum score.}
  \vspace{-0.3cm}
  \label{fig:per}
\end{figure} 
\subsection{Comparative Study}
Fig. \ref{fig:per} illustrates the performance of various methods in the BARN challenge. The green dashed box represents the performance of TEB and the current automatic tuning method, while the blue dashed box highlights the top three methods in the BARN challenge. The red dashed line denotes our method. It is evident that our approach significantly raises the performance ceiling of the automatic tuning algorithm and achieves first place in the BARN challenge.



\begin{figure*}[t]
  \centering
  \includegraphics[width=1\textwidth,height=0.6\textheight]{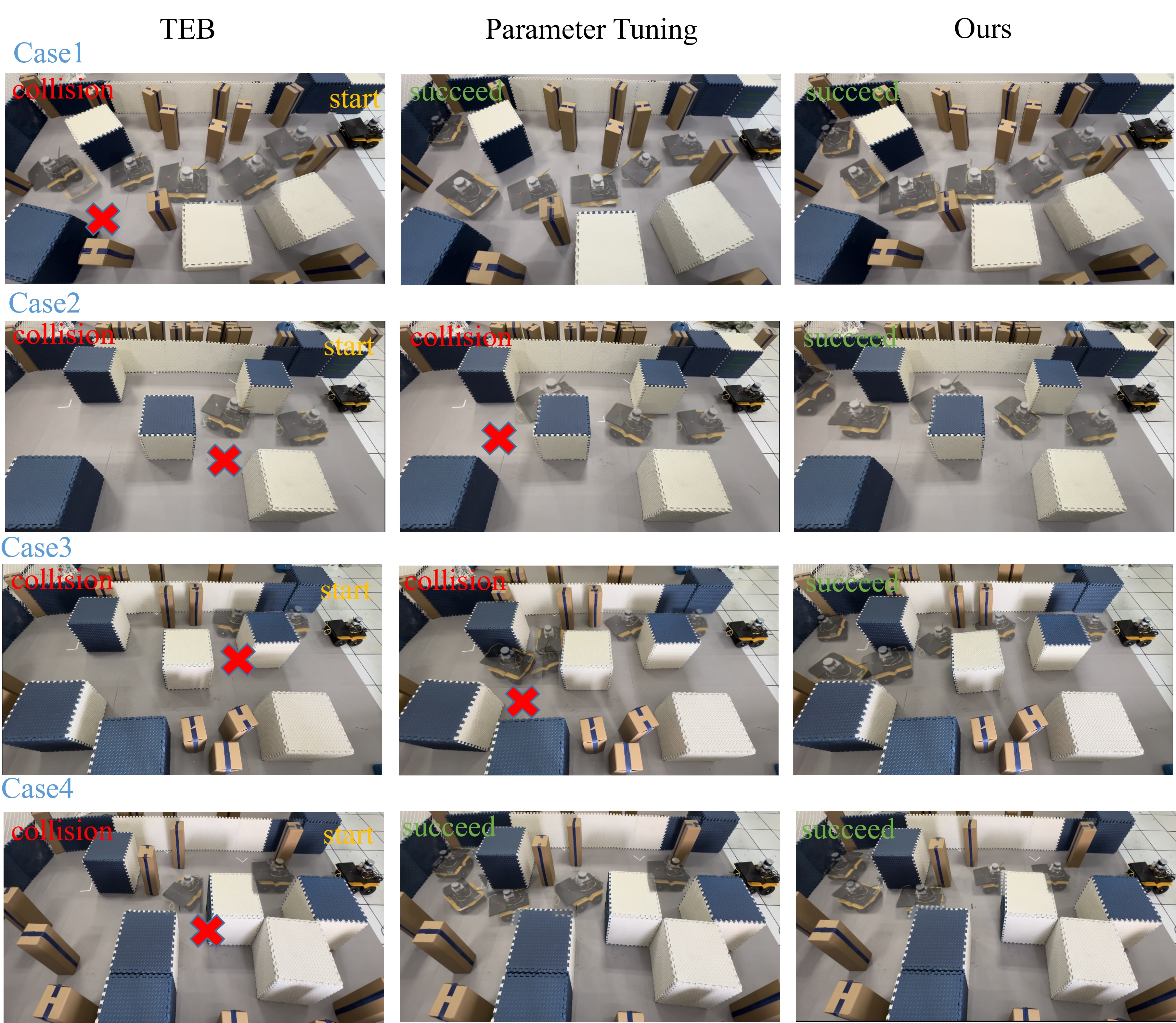} 
  \caption
  {\textcolor{black}Illustration of the real-world experiments 1-4.}
  \label{fig:real}
  \vspace{-0.3cm}
\end{figure*} 

\begin{table}[t]
\centering
\caption{Illustration of the real-world experiments.}
\setlength{\tabcolsep}{3pt} 
\renewcommand{\arraystretch}{1.2} 
\begin{tabular}{ccccccccc}
\toprule
\multirow{2}{*}{\textbf{Case}} & \multicolumn{2}{c}{\textbf{Collision Time (s)↑}} & \multicolumn{2}{c}{\textbf{Success Time (s)↓}} & \multicolumn{3}{c}{\textbf{Tracking Error (m)}↓} \\ 
\cmidrule(lr){2-3} \cmidrule(lr){4-5} \cmidrule(lr){6-8}
                          & \textbf{TEB}   & \textbf{PT+FC}  & \textbf{PT+FC}   & \textbf{Ours}  & \textbf{TEB}   & \textbf{PT+FC}  & \textbf{Ours}  \\ 
\midrule
1        & 6.3   & -                  & 10.7     & \textbf{8.4}                 & 0.045   & 0.037                   & \textbf{0.026}    \\
2        & 3.5   & \textbf{6.1}                  & -     & \textbf{8.2}                 & 0.046   & 0.044                   & \textbf{0.025}    \\
3       & 2.3   & \textbf{6.5}                  & -     & \textbf{9.4}                 & 0.045   & 0.048                   & \textbf{0.022}    \\
4     &  4.5   & -                  & 9.2     & \textbf{7.5}                 & 0.053   & 0.052                   & \textbf{0.034}    \\
5        & \textbf{8.7}   & 8.4                  & -     & \textbf{15.6}                 & 0.053   & 0.052                  & \textbf{0.028}    \\
6        & 6.1   & -                  & 13.2     & \textbf{11.7}                & 0.061   & 0.042                   & \textbf{0.033}    \\
7        & 4.4   & \textbf{6.7}                 & -     & \textbf{12.2}                 & 0.076   & 0.054                   & \textbf{0.032}    \\
8        & 4.8   & \textbf{5.2}                  & -     & \textbf{11.3}                 & 0.060   & 0.045                   & \textbf{0.034}    \\
\bottomrule
\label{tab:1}
\end{tabular}
\vspace{-0.2cm}
\end{table}

\subsection{Real-World Experiments}
We validate the algorithm's effectiveness through real-world experiments conducted on a physical Jackal robot. These experiments are performed in four distinct indoor environments and four corridor environments, which differ from any environments in the BRAN challenge. The 3D LiDAR data from a Velodyne sensor are converted into 2D data for use in these tests.
In each scenario, the start and end points are fixed, and three trials are conducted for each environment. The trajectory information is shown in Fig. \ref{fig:real}, where it is clear that the default TEB algorithm fails in all cases. The parameter tuning method (with a feedforward controller) is able to complete some of the cases, and with later collision times compared to the default TEB. This indicates that while parameter tuning alone provides some improvement, it is insufficient to achieve reliable success across all environments. Our proposed method, by contrast, demonstrates superior performance by successfully completing
all tasks with a 100\% success rate, the smallest tracking error and the shortest completion time. The detailed experimental results are shown in Tab. \ref{tab:1}.  


\section{CONCLUSIONS}
In this work, we propose a hierarchical architecture with low-frequency parameter tuning, mid-frequency planning, and high-frequency control. To mitigate the tracking error caused by pure feedforward control, we introduce an RL-based controller. Finally, we propose an iterative training framework from the
perspective of the hierarchical architecture, which alternates enhance the performance of the high-level parameter tuning and low-level control. Extensive simulations and real-world experiments are conducted to demonstrate the effectiveness of the proposed method.
\newpage

\bibliography{reference.bib}

\end{document}